\documentclass[conference]{IEEEtran}
\IEEEoverridecommandlockouts
\usepackage{cite}
\usepackage{amsmath,amssymb,amsfonts}
\usepackage{algorithmic}
\usepackage{graphicx}
\usepackage{textcomp}
\usepackage{xcolor}
\usepackage{multirow}
\usepackage{subcaption}
\usepackage{csquotes}

\begin{document}

\title{Joint Temporal Pooling for Improving Skeleton-based Action Recognition\\
}

\author{\IEEEauthorblockN{Shanaka Ramesh Gunasekara\textsuperscript{1} , Wanqing Li\textsuperscript{2},  Jack Yang\textsuperscript{3}, Philip Ogunbona \textsuperscript{4}}
\IEEEauthorblockA{\textit{Advanced Multimedia Research Lab} \\
\textit{University of Wollongong}\\
Australia\\
\textsuperscript{1}srg079@uowmail.edu.au, \textsuperscript{2}wanqing@uow.edu.au, \textsuperscript{3} jiey@uow.edu.au, \textsuperscript{4}philipo@uow.edu.au }}


\maketitle

\begin{abstract}
In skeleton-based human action recognition, temporal pooling is a critical step for capturing spatiotemporal relationship of joint dynamics. Conventional pooling methods overlook the preservation of motion information and treat each frame equally. However, in an action sequence, only a few segments of frames carry discriminative information related to the action. This paper presents a novel Joint Motion Adaptive Temporal Pooling (JMAP) method for improving skeleton-based action recognition. Two variants of JMAP, frame-wise pooling and joint-wise pooling, are introduced. The efficacy of JMAP has been validated through experiments on the popular NTU RGB+D 120 and PKU-MMD datasets.
\end{abstract}


\section{Introduction}

Skeleton-based human action recognition, which focuses on predicting actions from skeleton sequences, has emerged as a dominant modality in human action recognition. This is attributed to the skeleton's ability to provide abstract information about joint movements and their insensitivity to the environment. The skeleton is composed of 2D or 3D coordinates of joints.

One of the most popular and successful approaches to skeleton-based action recognition is based on Graph Convolutional Neural Networks (GCN) \cite{Chen2021a},  which model joint movements as a series of isomorphic graphs in the temporal dimension,  extract spatial features at each time step using graph convolutions, and perform temporal modelling using recurrent units \cite{Shuai01} or 1D convolutions \cite{Yan01, Maosen01,  Yong0123, Si0000, Cheng2017}. The common practice of temporal modelling is utilising multi-scale temporal convolution layers and temporal pooling such as stride convolution, averaging, and interpolation to reduce temporal dimensionality to preserve important temporal information \cite{Chen2021a, Liu6676}. 

\begin{figure*}[h!]
	\centering
	\includegraphics[scale=0.3]{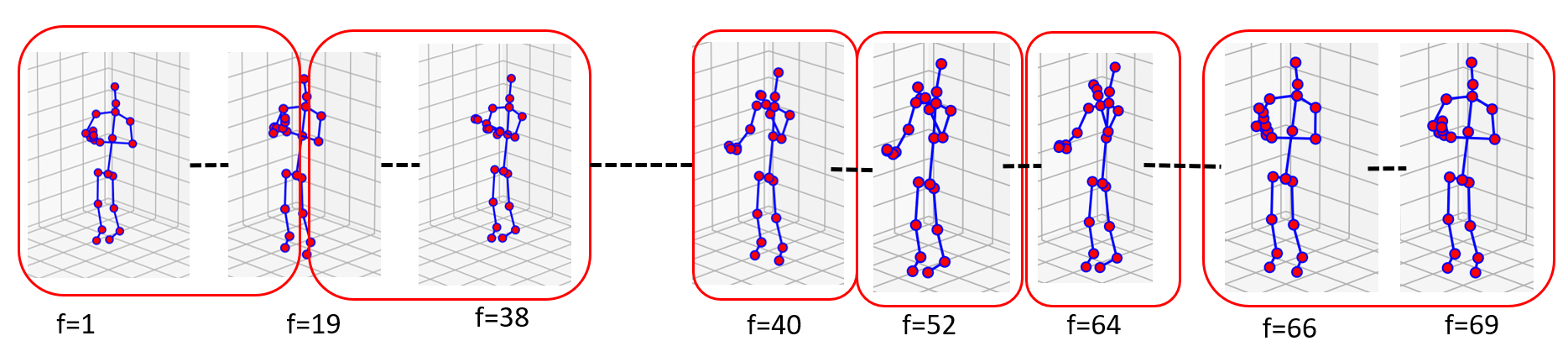}
    \caption{In a sequence of Drinking water action,  the person does not perform the action until f=38. At f=40 they start and stop at f=64 followed by a few frames which are not related much to the action. In conventional pooling methods, a constant pooling window size is applied across the sequence. But the proposed joint motion intensity adaptive pooling module changes the pooling window w.r.t the motion information and  generates pooling windows as in the figure (red boxes) to have wider windows to static segments and thinner windows to dynamic segments.}
	\label{fig00}
\end{figure*}

However, within an action sequence, only a few segments of frames often contain discriminative information that is relevant to the particular action.
For example, the first set of frames belonging to \enquote{Drinking Water} in Fig. \ref{fig00} exhibits a standing pose, which could also be shared by different actions such as \enquote{Eat a meal/snack}, \enquote{Brush your teeth}, and \enquote{Put on glasses}. Consequently, by following the conventional pooling process that treats all frames equally, the valuable features extracted from discriminative frames may be overshadowed by features derived from the remaining frames.

For robust action recognition, an ideal algorithm should look beyond local joint connectivity and exploit the complex cross-space-time cross-joint relationships for better recognition. Many existing works, use factorized spatial-only  and temporal-only convolution modules to capture temporal dynamics.  However,  joints are often not simultaneously active. For instance, in the action of \enquote{Counting money}, the wrist, elbow, and shoulder joints initiate the movement, followed by the wrist and finger joints to complete the action.  These joints are active  at distinct temporal locations within the sequence, necessitating a larger receptive field to effectively capture and model the dependencies among joints over time. Consequently, such actions are hard to recognize  even using multi-scale modelling. 

To alleviate the above limitations, a novel action recognition method is proposed  via the adaptive temporal pooling strategy based on joint motion intensities,  which is inspired by the work in \cite{Zhi1513,li2022}. A motion-sensitive frame selection method was discussed in \cite{Zhi1513} for RGB video sequences. In this work, that idea is extended as a joint temporal pooling for skeleton sequences by using joint motions. Two variants, frame-wise and joint-wise,  of the strategy, are introduced. In frame-wise  pooling, pooling is applied to the frame sequence. 
In the joint-wise approach, pooling is applied  to individual joints based on their motion intensity. This method  aligns discriminative joint-frame segments in the temporal domain to enable more effective cross-joint temporal  modelling using  a small receptive field. The proposed joint motion adaptive
temporal pooling (JMAP) can be seamlessly integrated into various action recognition baselines. Overall, the main contribution of the work is summarized as follows; 
\begin{enumerate}
    \item A joint motion intensity adaptive pooling layer is proposed to focus on the discriminative segments of action sequences for effective temporal pooling. 
    \item The proposed adaptive pooling module can be frame-wise as well as joint-wise to improve the cross-space-time and cross-joint modelling.
    \item Extensive experiments were conducted on  the NTU RGB+D 120, and  PKUMMD  datasets and results have validated the proposed JMAP with noticeable performance improvement.
\end{enumerate}

\section{Related work}

In this section, key works related to GCN-based skeleton action recognition and temporal pooling are reviewed.

\subsubsection*{\textbf{GCN-based Skeleton based action recognition}}

With the advance of Graph Convolutional Networks (GCNs), convolution operation was extended to non-Euclidean data like graphs and the model proposed in \cite{Kipf2017} has gained significant popularity. In the field of skeleton-based action recognition, recent models built on GCNs \cite{wang01,wang02},  have emerged as dominant approaches.   Yan et al. \cite{Yan01} introduced the use of graph networks to model skeleton dynamics with ST-GCN. Recent research has showcased the effectiveness of these networks in capturing intrinsic relationships by leveraging data-driven adaptive learning of joint correlations and employing channel-wise joint topology modelling. \cite{Shi2019,cong01,lei02,Chen2021a,Miao01,Chuankun01}. GCN models mainly utilize graph convolution for spatial modelling. Various approaches have been explored in temporal modelling, including recurrent units \cite{Yong0123,Si0000} and 1D convolutions \cite{Cheng2017}. More advanced techniques involve multi-scale temporal convolution layers \cite{Liu6676,Chen2021a} that incorporate temporal pooling techniques such as stride convolution and dilation.   However, simultaneously acquiring local and global variations of temporal features are restricted in these models. Besides, existing methods suffer limitations of ineffective selection of discriminative frame segments with motion information.  Moreover, these methods do not explicitly address the expression of cross-joint correlations across different time intervals. GCNs were chosen as the experimental baselines for validating  the proposed  JMAP method.


\subsubsection*{\textbf{Temporal pooling and frame selection}}
In action sequence modelling, conventional temporal pooling techniques often lead to poor performance, and alternative methods such as convolution-based \cite{Sun2015} and self-attention-based pooling \cite{Liu1362,Zhang1357} have been proposed. Recently, self-attention-based approaches have emerged as possible alternatives to replace conventional temporal pooling methods. Selective feature compression (SFC) \cite{Liu1362}  maps the long video sequence into a  small representation in the feature space by using an attention-based mechanism. In the VidTr model \cite{Zhang1357}, temporal downsampling was conducted by utilizing an attention matrix to select the top K highly activated set of temporal frames, while discarding frames with uniform attention scores. However, these methods rely heavily on dot product attention.

Frame selection or frame sampling is another common approach used to control temporal redundancy by sampling frames at a fixed stride,  randomly, or based on a predefined algorithm. In Temporal Segment Network (TSN) \cite{Wang1608} proposed a sampling method that selects snippets from uniformly segmented clips within the sequence. However, TSN treats each frame equally and only aims to remove the redundancy between adjacency frames. Similarly, Adaframe \cite{Wu1127} introduced a policy gradient-trained LSTM with a global memory component to dynamically search and select frames over time.  However, these methods discard the motion information  within the adjacent frames. In contrast,  SCSampler \cite{Bruno110}, was proposed to select salient clips, utilizing a lightweight CNN as the section network based on salience score.  They also incorporated audio channels as an additional input modality for network training.  One disadvantage of these selection modules is their reliance on separate policy networks, which introduce additional learnable parameters. Additionally, training such models requires a large amount of data and significantly increases training time.  Moreover, none of these methods considers motion information during the sampling process. MGSampler \cite{Zhi1513}introduces a motion-preserving temporal pooling strategy in RGB video clips, which defines an adaptive pooling window based on pixel-wise motion intensity. By dividing the sequence into segments according to the motion intensity curve, MGSampler selects a random frame from each segment to obtain the pooled output sequence. This idea is extended into a learnable pooling method by introducing a pooling matrix in \cite{li2022}. Building upon these influences, joint motion adaptive temporal pooling is introduced for skeleton action recognition, focusing on motion information during pooling. Importantly, this approach brings about negligible computational overhead while enhancing the accuracy of the recognition.

\begin{figure*}[h!]
	\centering
	\includegraphics[scale=0.5]{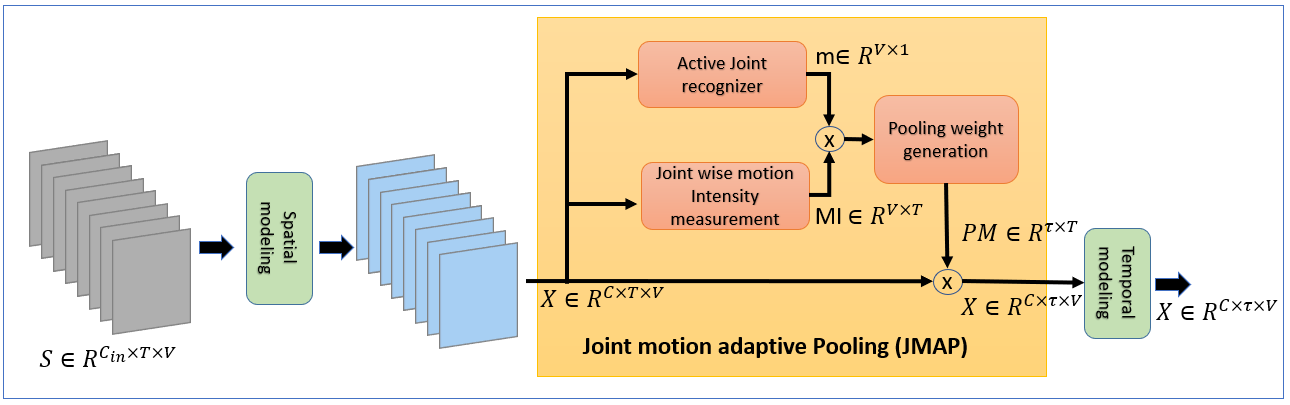}
	\caption{ The pipeline for joint motion adaptive pooling module}
	\label{fig:01}
\end{figure*}


%


\section{Proposed Method}

The proposed JMAP (Joint Motion Adaptive Pooling) involves extracting discriminative segments from the skeleton sequence based on joint motion intensities. First, a general overview of the problem is provided. The subsequent section presents a detailed description of joint motion calculation and elaborates on the selection of discriminative frame segments, which ultimately leads to the application of an adaptive pooling window. Fig. \ref{fig:01} provides an overview of the proposed JMAP.

\subsection{Problem formation}

For a given skeleton sequence input $X \in R^{C \times T \times V}$, where $C$ is the number of channels, $T$ is the temporal dimension, and $V$ represents the number of joints. The pooling task involves scaling the temporal dimension by a factor of $\theta$ as $\tau = T / \theta$. The resulting pooled output is denoted as $X' \in R^{C \times \tau \times V}$. Conventional temporal pooling methods in action recognition typically employ uniform pooling windows across the temporal domain, treating all frames equally. However, this approach fails to effectively preserve motion information within consecutive frames. In contrast, JMAP dynamically defines the pooling window based on the motion of joints at different locations within the temporal domain, thus effectively preserving crucial motion while pooling, as illustrated in Fig \ref{fig00}. The joint motion representation relies on the temporal differences between consecutive frames for each joint, while the selection of discriminative segments is based on the cumulative joint motion intensity across the sequence.
 

\subsection{Joint motion adaptive pooling}

In a skeleton sequence, each frame typically represents a static pose at a specific time point. The motion of the skeleton is defined by the feature-level joint-wise temporal difference between two such poses. This joint difference between frames focuses primarily on motion-specific features, which are crucial for action recognition tasks, while suppressing the static joints. The proposed JMAP utilizes motion features for adaptive pooling and consists of two main parts: (a) joint motion intensity measurement and (b) adaptive pooling window selection.


\subsubsection{\textbf{Joint-based motion intensity measurement}}

\hfill \break
In JMAP, the joint motion intensity is calculated by considering the magnitude of the joint motion in consecutive frames, while disregarding the direction. The motion intensity is computed by taking the joint-wise feature-level difference between consecutive frames and accumulating it over the channel dimension. Formally, for a given input $X \in R^{C \times T \times V}$, the motion intensity for the joint $v \in V$ is computed as follows:

\begin{align}
    MI_{v,t} = \dfrac{1}{C} \Sigma_{c=1}^{C} |X(t,v,c) - X((t-1),v,c)| , t \in \{ 2,3,...,T\},
    \label{eq:1}
 \end{align}
where $MI_{v,t} \in \mathcal{R}^{V \times T}$ is the joint motion signals. 
Note that not every joint is involved in performing every action. 


 \hfill \break
 \textbf{Active Joint: } Active joints are identified as joints that significantly contribute to performing a specific action, recognizing that not every joint plays a significant role, and the number of active joints can vary across different actions. Therefore, the active joints for each sequence should be adaptively selected based on the criteria for calculating motion intensity used in pooling.

The kinematic and kinetic information of the skeleton change as the joints' positions in 3D space undergo temporal changes. Based on the assumption that the majority of the active joints contribute to the overall energy change of the whole body, the active joint selection criteria are defined as follows.

The kinetic energy $E_v^{kin}$ and potential energy $E_v^{pot}$ over the sequence are calculated by considering joint-wise dynamics.
Thus, 
\begin{align}
    E_v^{kin} = \Sigma_{t=2}^T\dfrac{1}{C} \Sigma_{c=1}^{C} (X(t,v,c) - X((t-1),v,c))^2,
    \label{eq:2}
 \end{align}
and 
 \begin{align}
    E_v^{pot} = |\Sigma_{t=2}^T\dfrac{1}{C} \Sigma_{c=1}^{C} (X(t,v,c) - X(1,v,c))|.
    \label{eq:3}
 \end{align}

When calculating the potential energy, the first frame is taken as the reference zero energy pose, and only the absolute value is computed since the magnitude of the energy change is what matters. The total energy of a joint is defined as the summation of these two terms and then scaled to the range  [0, 1] by applying a two-step normalization process relative to the maximum energy.  Total normalized energy is computed as:

\begin{align}
    E_{v-norm}  = \dfrac{E_v}{max (E_V)} ,
    \label{eq:4}
 \end{align}

where $E_v$ is the energy of joints $v$ and $E_V$ is a vector that includes the energy of all joints as $E_V = [E_1,E_2,...,E_v]$. 

First, the individual energy terms are normalized, followed by the normalization of the total energy of the joint $v$, denoted by $E_v^{tot} = E^{kin}_{v-norm}+E^{pot}_{v-norm}$. It is observed that certain joints may initially exhibit high potential energy change due to pose changes but subsequently become inactive and remain at rest. By utilizing the two-step normalization process, the total joint energy is effectively mapped within the range  [0, 1]. This normalization preserves the intrinsic characteristics of the energy values without introducing any disturbances or alterations.

The active joint contribution to the total energy change is estimated by considering the mean $\mu$ and standard deviation $\sigma$ of the joint-wise energy distribution:

\begin{align}
    \mu = \dfrac{\Sigma_{v=1}^V {E_v^{tot}}}{V} ,
    \label{eq:5}
 \end{align}

  \begin{align}
    \sigma  = \sqrt{ \dfrac{{\Sigma_{v=1}^V ( E_v^{tot} - \mu) } } {V-1}}.
    \label{eq:6}
 \end{align}

The set of active joints is selected such that, $ V_{active}  = {[V_{active}; E^{tot}_{v-norm} \geq \mu - \alpha \times  \sigma ]}$ and $\alpha$ is a control variable to select the number of active joints which is dynamically adjusted based on the energy values observed in each input $X$.

\subsubsection{\textbf{Adaptive pooling window }}
\hfill \break
The subsequent step involves the selection of an adaptive pooling window size which is a function of the obtained prominent joint motion signals over time, $f(MI_{v,t})$. The window size is determined by examining the entire joint motion throughout the sequence, leveraging the cumulative joint motion $CI_{v,t}$ as introduced in prior works \cite{Zhi1513,li2022}. Prior to analyzing the cumulative joint motion, a normalization step is introduced to enhance the smoothness of the $CI_{v,t}$ curve. The purpose of this normalization is to amplify high motion intensity values while attenuating low motion intensity values, allowing for better discrimination between discriminative frames and others.



\begin{align}
   & \Tilde{MI_{v,t}} = \dfrac{tanh({MI_{v,t}})}{\Sigma_{t=1}^T tanh({MI_{v,t}}) + \epsilon_2} ; t>0,
   \label{eq:9}
 \end{align}
where $\Tilde{MI_t}$  is the normalized motion intensity and initial value set as $\Tilde{MI_t} = 0 ; t=0 $
 \begin{align}
   CI_{v,t} = \Sigma_{t'\leq t} \Tilde{MI_{v,t}}.
   \label{eq:10}
 \end{align}
 
In order to address the issue of infinity values during backpropagation caused by zero values in the input, a small $\epsilon_2$ is introduced. The gradient of the cumulative intensity curve reflects the motion intensity at each time step. A lower gradient indicates a static position with non-informative frames in the sequence, while a higher gradient represents motion-informative frame segments more details are verified in the experimental session.



To obtain $\tau$ frames after pooling, the y-axis of the curve $CI_{v,t}$ is divided into equal $\tau = T/\theta$ segments, where $\theta$ is the scaling factor. The adaptive pooling windows are then defined as the intersection points of the x-axis with the boundaries of the curve. The size of the pooling windows depends solely on the curvature of the curve, enabling a smaller window for discriminative frame segments, as illustrated in Fig. \ref{fig4} and Fig.\ref{fig5}.
The frames within each pooling window are compressed using learnable weights, as defined in previous work \cite{li2022}, with a learnable pooling matrix $P_m \in \mathbb{R}^{\tau \times T}$ used for pooling along the temporal dimension, denoted as $X' = P_m X$, where $X' \in \mathbb{R}^{C \times \tau \times V}$. The elements of the matrix $P_m$ are computed as follows:

\begin{align}
    P_m(i,j) = \dfrac{1}{(\dfrac{CI_j - m_i}{0.5w})^{2\Gamma}+1},
    \label{eq:11}
    \end{align} 

where parameter $\Gamma$ serves as a temperature term that regulates the smoothness of the rectangular smoothing function, while $w$ represents the width of the segments on the y-axis. Higher values of $\Gamma$ encourage standard temporal average pooling. The pooled sequence is rich with motion information as it preserves the significant motion information in the latent space.

\subsection{Joint wise and frame wise pooling}

The proposed JMAP can be implemented in  two pooling strategies: joint-wise pooling and frame-wise pooling. In joint-wise pooling, the motion intensities of each joint are considered independently. The proposed JMAP is applied to the active joints, while the rest are pooled using a constant pooling window.
In frame-wise pooling, the $V$-dimensional joint motion signal $MI_{v,t}$ is transformed into a one-dimensional motion signal that represents the motion magnitude at each frame by accumulating the motion intensities over the joint dimension.

\begin{align}
    MI_t =\dfrac{1}{\Sigma m} \Sigma_{v=1}^V (m \cdot MI_{v,t}),
    \label{eq:8}
 \end{align}

where $m \in \mathcal{R}^{V\times 1}$ is a binary mask and 1 is given for active joints and 0 is for inactive joints. Then  using the obtained cumulative frame motion intensity curve, the proposed JMAP is applied.


\section{Experiments}

 \subsection{Datasets}
The proposed method is evaluated  on two commonly used large scale skeleton datasets. 
 \subsubsection*{\textbf{NTU RGB+D 120}}

 NTU RGB+D 120 \cite{Liu1905} is an extension of the NTU RGB+D data set having a total of 113,945 skeleton sequences over 120 different classes performed by 106 participants making it the largest and most challenging skeleton-based action recognition dataset. Each skeleton consists of 25 3D body joints.  The evaluation is based on two protocols 1. Cross-subject (X-sub): Training set comprises samples from 56 subjects; Test set comprises samples from 50 subjects. 2. Cross-set (X-set): Training set comprises samples from half of the camera setup; Test set comprises the other half.
 
 \subsubsection*{\textbf{PKUMMD}}
The PKU-MMD dataset \cite{Liu1703},  is another recently introduced, popular skeleton-based large dataset, containing 1076 long untrimmed videos  of skeleton sequences. Here, 51 actions are performed by 66 distinct subjects in three camera views and 3D space coordinates of 25 joints are provided. We retrieved 21,545 valid action sequences from the untrimmed videos using the algorithm developed in \cite{Yu3199}.  Similarly to the NTU RGB+D dataset, this dataset is also evaluated on X-sub and X-view  protocols. 

\subsection{Implementation and model architecture}
 CTR-GCN \cite{Chen2021a}, ST-GCN \cite{Yan01}, and 2s-AGCN \cite{lei02} were selected as the backbone network for the implementation of the proposed joint motion adaptive layer. The CTR-GCN utilizes a multi-scale temporal module, while ST-GCN and 2s-AGCN employ temporal convolution layers for temporal modelling. In these networks, the stride convolution pooling, which performs temporal pooling and scales the temporal dimension by half, is replaced with the proposed JMAP module. The value of $\alpha$ is set to 0.1 for selecting active joints, and $\Gamma$ is set to 5 in Equation \ref{eq:11}. These parameters were obtained empirically. The experiments were conducted on RTX 3090Ti GPU server. During the training of the networks, the respective training parameters specified in their original repositories were used. Both the NTU RGB+D 120 and PKUMMD datasets were pre-processed following the method described in \cite{Zhang1904}.

\subsection{Performance of JMAP with different backbones }
The proposed pooling module offers the advantage of being easily integrated into various action recognition models. To ensure a fair comparison, we applied the same pre-processing steps and utilized only the joint information from the NTU RGB+D 120 dataset. In our approach,  the layers responsible for temporal pooling in the original network were replaced with the proposed JMAP module, while keeping the remaining layers unchanged, following the default configuration.


\begin{table}[htbp]
\caption{Performance of our module by comparing Top-1 accuracy using different GCN-backbones on NTU-RGB+D 120 dataset with X-Sub and X-Set settings using \textit{single} joint input modality}
\begin{center}
\begin{tabular}{l l l}
\hline 
\multirow{2}{*}{Model archtecture} & \multicolumn{2}{l}{NTU RGB 120}  \\
                                   & X sub (\%)& X set (\%)                  \\
 \hline \hline
ST-GCN                          & 82.93   & 84.83                      \\
+ JMAP frame wise               & 83.14     & 85.12                      \\
+ JMAP joint wise               & 83.57     & 85.37                     \\
\hline
2S-AGCN                            & 84.21   & 85.61                       \\
+ JMAP frame wise                  &  84.33    &   85.93                    \\
+ JMAP joint wise                  & 84.54   & 86.14                       \\
\hline
CTR-GCN                         & 84.46    & 86.12                      \\
+ JMAP frame wise               & 84.76    & 86.34                     \\
+ JMAP joint wise               & 85.23    & 86.87                       \\
\hline

\end{tabular}
\label{tab1}
\end{center}
\end{table}

The performances of all baseline models were observed to improve with the proposed JMAP in comparison to models utilizing conventional temporal pooling modules as summarized in Table \ref{tab1}.


\subsection{Performance on Ambiguous Actions}
Ambiguous actions are categorized according to the CTR-GCN baseline on NTU RGB+D 120 X sub. Actions whose accuracy of recognition is between 60\% to 75\%  are considered as hard actions, and actions with an accuracy below 60\% are considered extremely hard actions. The improvement compared to the baseline is then assessed. With joint-wise JMAP, out of six extremely hard actions, four action classes showed improvement. Among the 20 hard actions, 14 actions demonstrated improvement, as illustrated in Figure 3. However, there were six actions that experienced a drop in accuracy. These actions are performed using only a few specific joints, and their scope of movement may also be quite limited. To improve the accuracy of such actions, it is necessary to enhance the active joint selection criteria to carefully choose only the joints that actively contribute to the action while having a very small motion scope.
     \begin{figure}[htbp]
	\centering{\includegraphics[scale=0.5]{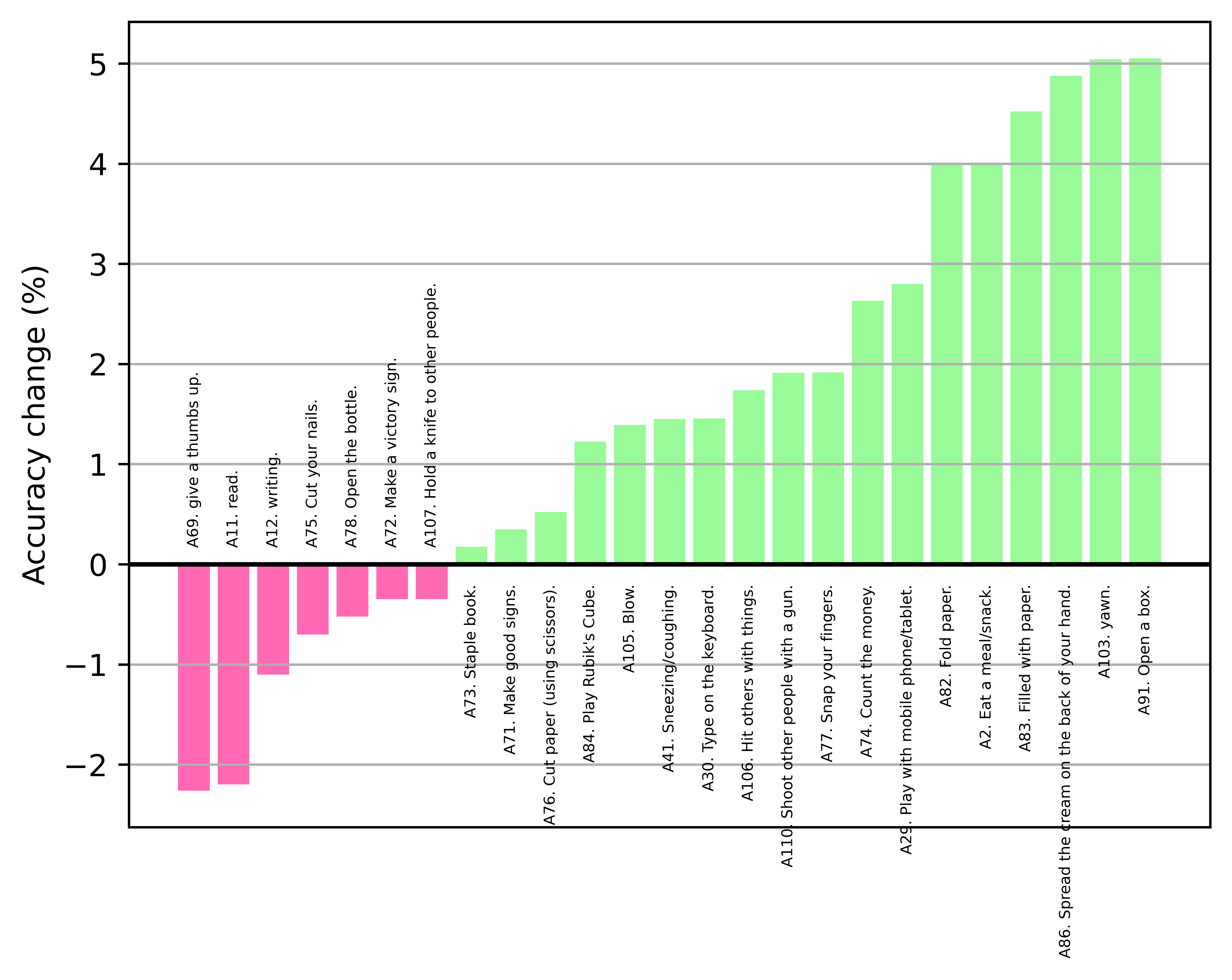}}
	\caption{ Accuracy change between the CTR-GCN and with JMAP for hard and extreme hard action group.}
	\label{fig:03}
        \end{figure}

\subsection{Comparison with the State-of-the-Art}
A comparison with state-of-the-art methods is presented, using NTU RGB+D 120, and PKUMMD datasets to demonstrate the competitiveness of the proposed JMAP module. Table \ref{tab2} and Table \ref{tab3} present the quantitative comparison of the result. Most of the state-of-the-art methods use mutli-stream fusion to obtain the stated results \cite{Chen2021a}. In a similar manner, four modalities (joint, bone, joint motion, and bone motion) were fused for the results based on NTU RGB+D 120 dataset. The most recent reported action recognition result for the PKUMMD dataset is from \cite{Li1909}, where only the joint modality is used.
    
\begin{table}[htbp]
\caption{Performance comparison of skeleton-based action recognition mAP\%  on PKU MMD dataset with \textit{single} joint input modality}
\begin{center}
  \begin{tabular}{ l l l } 
\hline 
\textbf{Method} &  \textbf{X-sub (\%)} & \textbf{X-view (\%)} \\ [0.5ex] 

\hline \hline
Li et al. \cite{Li1704a} & 90.4& 93.7 \\
HCN \cite{Li1804} & 92.6 & 94.2\\
RF-Action \cite{Li1909} &  92.9& 94.4\\
\hline \hline
Ours (Frame wise JMPA) & 94.12 & 95.51\\
Ours (joint wise JMPA) & \textbf{94.43}  & \textbf{95.84}\\
\hline
  \end{tabular}

\label{tab2}
\end{center}
\end{table}

\begin{table}[ht]
\caption{Performance comparison of skeleton-based action recognition in top-1 accuracy on NTU-RGB+D 120 data set  with multi-stream fusing (Joint, Bone, Joint motion, Bone motion) } 
\centering 
\begin{tabular}{ l l l } 
\hline 
\textbf{Method} &  \textbf{X-sub (\%)} & \textbf{X-set (\%)} \\ [0.5ex] 

\hline \hline
2S-AGCN \cite{lei02} & 82.9& 84.9 \\
ST-GDN \cite{Peng2021} & 80.8 & 82.3\\
GAS-GCN \cite{Chan2020}  & 86.4& 88.0 \\ 
SGN \cite{Zhang1904}  & 79.2  & 81.5\\
shift-GCN \cite{Cheng2017} & 85.9 & 87.6 \\
MS-G3D \cite{Liu6676} & 86.9 & 88.4\\
MST-GCN \cite{Chen2021} & 87.5 & 88.8 \\
CTR-GCN \cite{Chen2021a} & 88.9 & 90.6 \\
\hline \hline
Ours (Frame wise JMPA) & 89.1 & 90.7\\
Ours (joint wise JMPA) & \textbf{89.4} & \textbf{90.9}\\
\hline


\end{tabular} 
\label{tab3} 
\end{table}
\subsection{Ablation study }
Ablation studies were conducted to evaluate different normalization functions and assess the effectiveness of the pooling layer at different network levels using the X-sub benchmark of the NTU RGB+D 120 dataset. The CTR-GCN model was used as the baseline for the experiments.

\subsubsection*{\textbf{The effect of the pooling layer at different levels of the network}}

 An experiment was conducted by changing the position of the proposed joint-wise JMAP in the network. The original CTR-GCN \cite{Chen2021a} model applies temporal pooling at layer 5 and layer 8, and observed the effectiveness of our proposed model by replacing these two pooling layers independently. Additionally, joint-wise JMAP was applied at layer 1. In this setup,  the baseline model's pooling at layer 5 is replaced while keeping the pooling at layer 8 unchanged. The accuracy was dropped in this configuration, as the motion signals contain a lot of noise at the beginning of the network, which disturbs the true motion patterns of the joints. The results of this experiment are presented in Table \ref{tab4}.

\begin{table}[htbp]
\caption{Ablation studies of the effectiveness of the joint-wise JMAP at different levels of the network on NTU-RGB+D 120 dataset under the X-Sub setting with the joint input modality. Top-1 accuracy is reported}
\begin{center}
\begin{tabular}{l l}
\hline
\textbf{Pooling layer at} & \textbf{\textit{Acc \%}} \\
\hline \hline
Baseline & 84.48  \\
\hline
layer 1 &  83.20   \\
layer 5 &  84.72   \\
layer 8 &   84.86  \\
layer 5 and 8 & \textbf{85.23} \\
\hline
\end{tabular}
\label{tab4}
\end{center}
\end{table}

\subsubsection*{\textbf{Effect of the normalization function}}

\begin{figure}
    \centering
    \begin{subfigure}[b]{0.45\columnwidth}
        \includegraphics[scale =0.2,width=\linewidth]{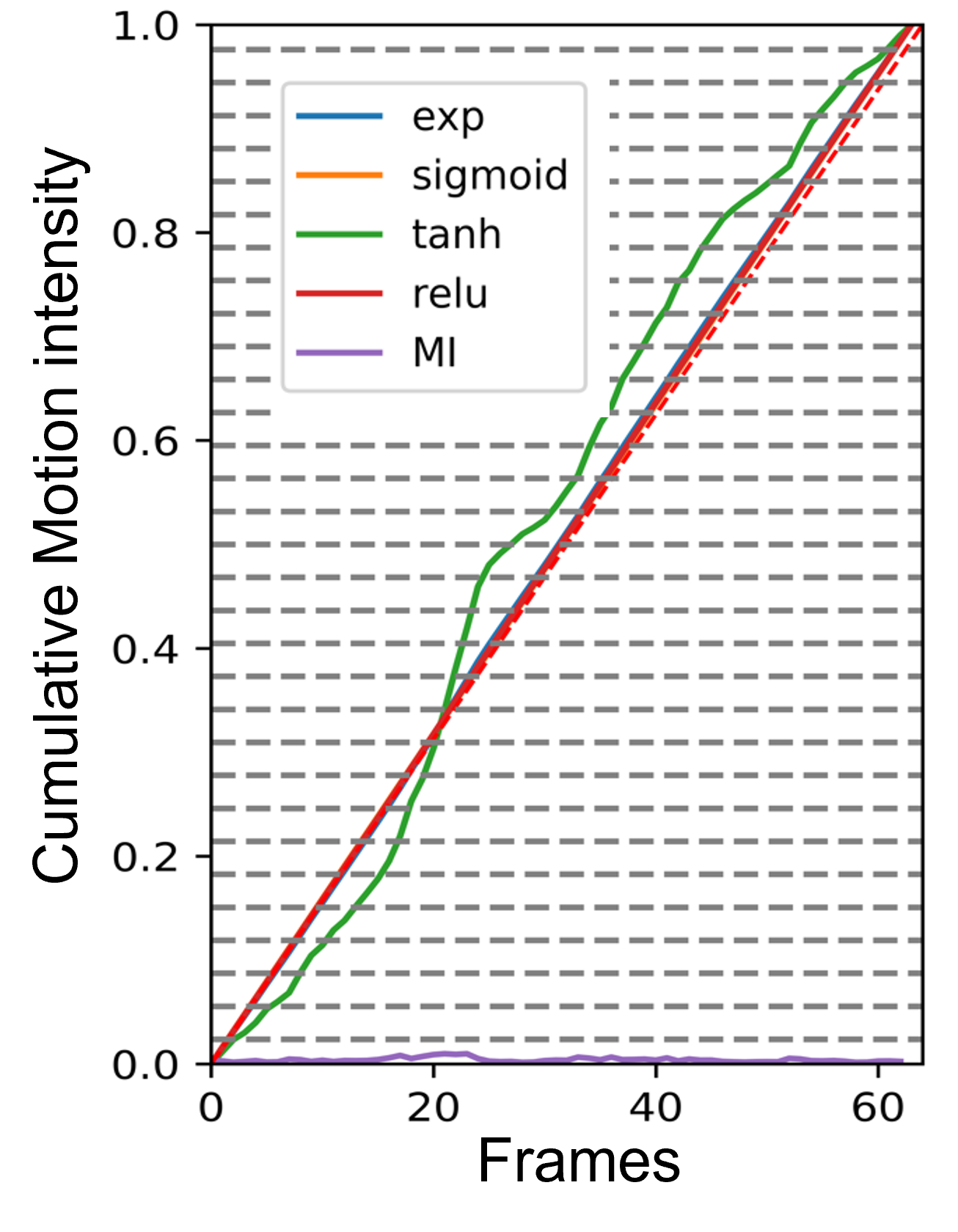}
        \caption{A4. Comb your hair }
        \label{fig:sub1}
    \end{subfigure}
    \hfill
    \begin{subfigure}[b]{0.45\columnwidth}
        \includegraphics[scale =0.2,width=\linewidth]{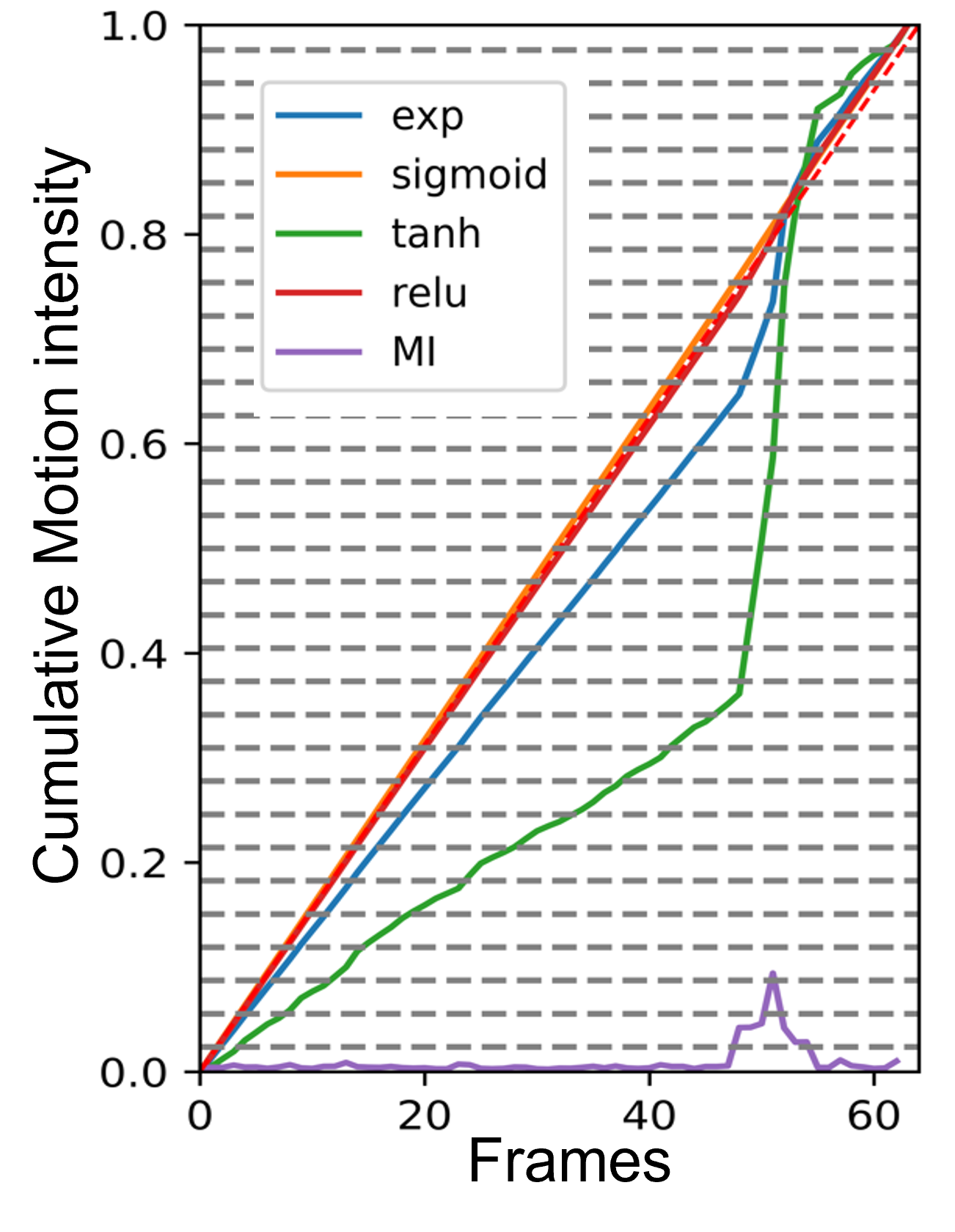}
        \caption{A68. Flick your hair}
        \label{fig:sub2}
    \end{subfigure}
    \caption{Cumulative Joint Motion Intensity (CJMI) curves with different normalization functions. The CJMI with $tanh$, gives the best curvature with high motion intensities and  while the rest of the functions generate curves more aligned with uniform pooling (red dashed line)}
    \label{fig4}
\end{figure}

The normalization function in Eq. \ref{eq:9} is employed to effectively distinguish motion-discriminative frames from other frames. The objective of this function is to generate a high gradient for the discriminative frame segments, thereby emphasizing their significance, while producing a low gradient for the rest of the frames in the $CT_t$ curve. The effectiveness of the normalization function is analysed on randomly selected samples belonging to different actions, as depicted in Fig. \ref{fig4}. For this experiment, frame-wise motion intensity was calculated using only selective active joints. It is evident from the visualization that the $tanh$ function exhibits better curvature as desired.

\begin{figure}
    \centering
    \begin{subfigure}[b]{0.45\columnwidth}
        \includegraphics[scale =0.2,width=\linewidth]{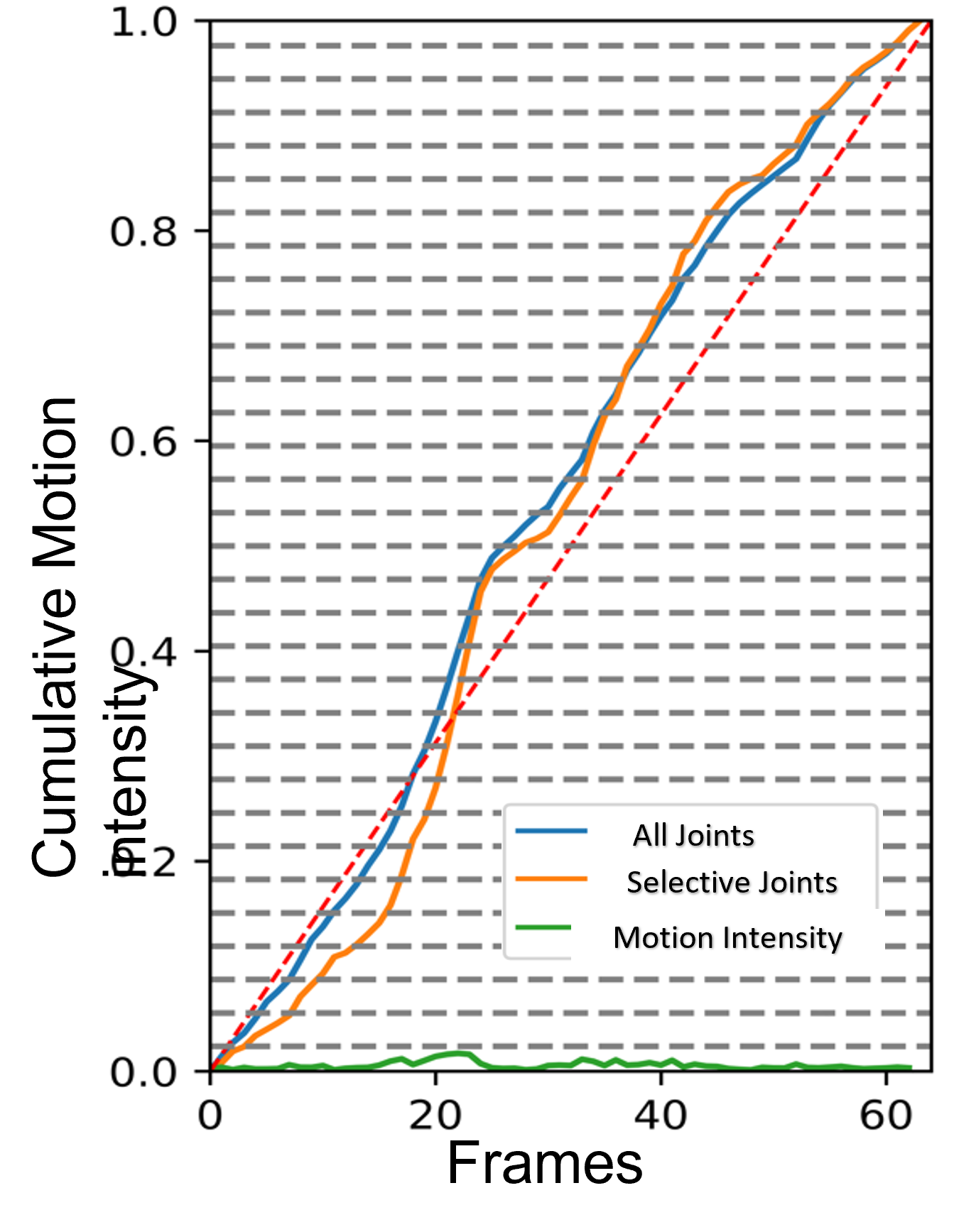}
        \caption{A4. Comb your hair}
        \label{fig:sub1}
    \end{subfigure}
    \hfill
    \begin{subfigure}[b]{0.45\columnwidth}
        \includegraphics[scale =0.2,width=\linewidth]{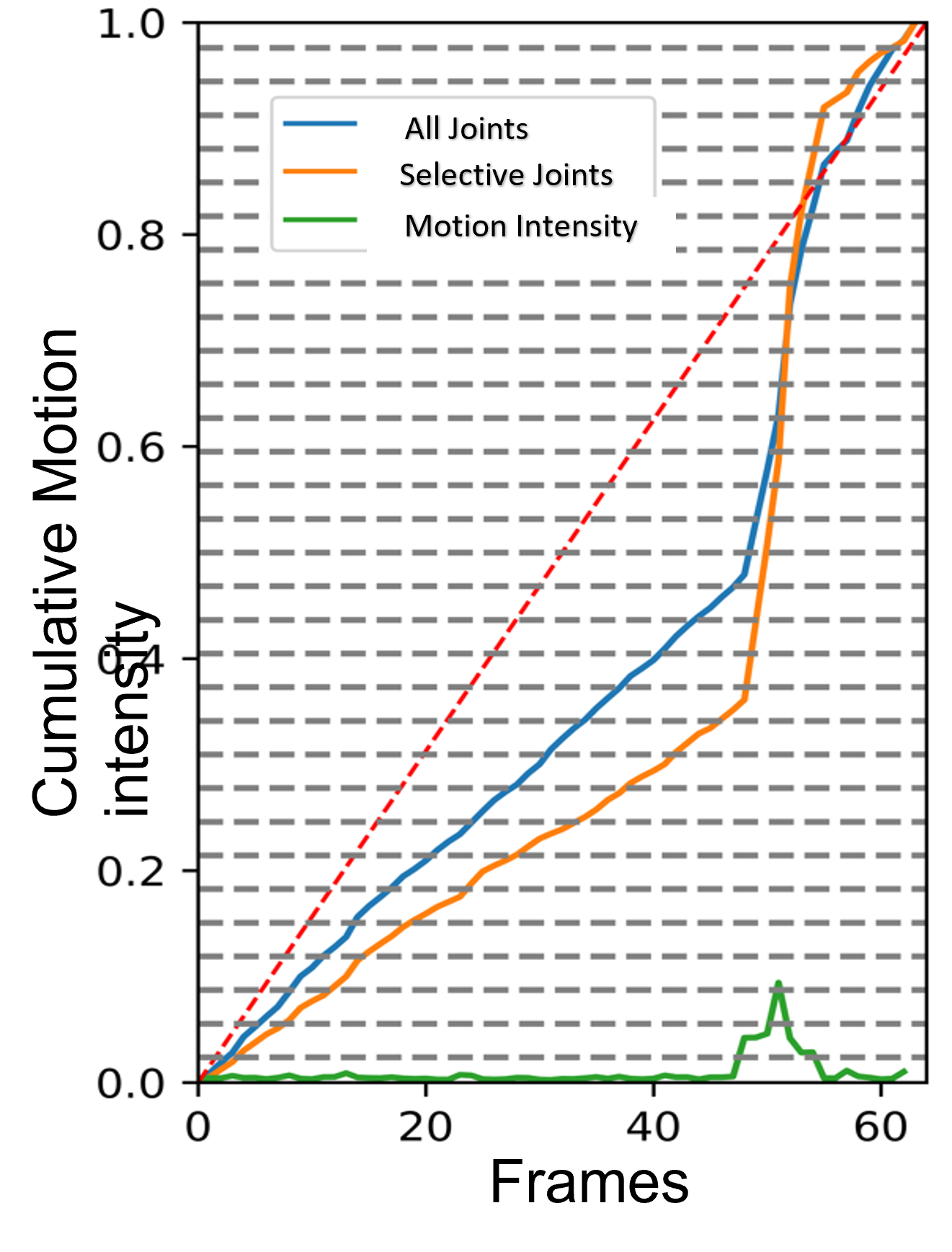}
        \caption{A68. Flick your hair}
        \label{fig:sub2}
    \end{subfigure}
    \caption{CJMI curve obtained using All joints and Active joints are presented. The CMJI with active joints shows a better curvature compared to All joints. The red dashed line belongs to uniform sampling}
    \label{fig5}
\end{figure}

At the same time, the significance of selecting active joints is also analyzed using a few samples. The pooling window size should be small for the discriminative frames which is determined by the gradient of the cumulative motion curve. Higher curvature of the curve gives a small window size in the pooling and with selective joints the curvature for the discriminative frames is higher compared to all joints which leads to a smaller window size for pooling to capture more details. This is illustrated in Fig.\ref{fig5}.

\begin{figure}[h]
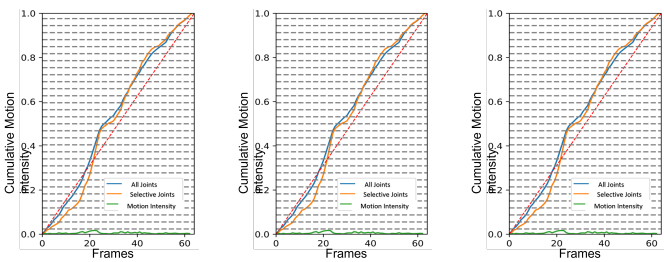

  \begin{subfigure}[b]{0.3\columnwidth} 
    \centering
    \includegraphics[scale =0.2,width=\linewidth]{fig113.png}
    \caption{ A}
    \label{fig:sub1}
  \end{subfigure}
    \hfill
  \begin{subfigure}[b]{0.3\columnwidth} 
    \centering
    \includegraphics[scale =0.2,width=\linewidth]{fig113.png} 
    \caption{B}
    \label{fig:sub2}
  \end{subfigure}
  \hfill
  \begin{subfigure}[b]{0.3\columnwidth} 
    \centering
    \includegraphics[scale =0.2,width=\linewidth]{fig113.png}  
    \caption{C}
    \label{fig:sub3}
  \end{subfigure}
  \caption{Main Figure with Subfigures}
  \label{fig:main}
\end{figure}


\section{Conclusion}
This paper presents a novel joint motion temporal pooling module for skeleton-based action recognition. The module is designed to prioritize motion-informative segments within the sequence during temporal pooling. The proposed method includes two variations: frame-wise pooling and joint-wise pooling. These variations dynamically define the pooling window to preserve motion information by utilizing active joint motion intensities from the action sequence. Extensive experiments validate the effectiveness of the proposed module in accurately distinguishing confusing categories and its compatibility with most action recognition backbones. The method outperforms state-of-the-art methods on two widely used benchmarks.

\bibliographystyle{unsrt}
\bibliography{bibliography}


\end{document}